\documentclass[acmsmall, review=false]{acmart}

\usepackage{hyperref}
\hypersetup{
    colorlinks,
    citecolor=red,
    filecolor=black,
    linkcolor=black,
    urlcolor=blue
}

\newcommand{\etal}{\textit{et al. }}

\newcommand*{\new}[1]{\textcolor{black}{#1}}

\usepackage{pifont}
\usepackage{amsmath} 
\usepackage{adjustbox} 

\newcommand{\cmark}{\ding{51}}%
\newcommand{\xmark}{\ding{55}}%

\usepackage[flushleft]{threeparttable}

\usepackage{makecell}

\AtBeginDocument{%
  \providecommand\BibTeX{{%
    \normalfont B\kern-0.5em{\scshape i\kern-0.25em b}\kern-0.8em\TeX}}}

\setcopyright{none}
\copyrightyear{2022}

\renewcommand\footnotetextcopyrightpermission[1]{}

\begin{document}

\title{Intelligence at the Extreme Edge: A Survey on Reformable TinyML}

\author{Visal Rajapakse}
\authornotemark[1]
\email{visalrajapakse@gmail.com}
\affiliation{
  \institution{University of Westminster}
  \city{London}
  \country{England}
}
\author{Ishan Karunanayake}
\authornotemark[2]
\email{ishan.karunanayake@unsw.edu.au}
\affiliation{
  \institution{University of New South Wales}
  \city{Sydney}
  \country{Australia}
}
\author{Nadeem Ahmed}
\authornotemark[3]
\email{nadeem.ahmed@unsw.edu.au}
\affiliation{
  \institution{University of New South Wales}
  \city{Sydney}
  \country{Australia}
}

\begin{abstract}
Tiny Machine Learning (TinyML) is an upsurging research field that proposes to democratize the use of Machine Learning and Deep Learning on highly energy-efficient frugal Microcontroller Units. Considering the general assumption that TinyML can only run inference, growing interest in the domain has led to work that makes them reformable, i.e., solutions that permit models to improve once deployed. This work presents a survey on reformable TinyML solutions with the proposal of a novel taxonomy. Here, the suitability of each hierarchical layer for reformability is discussed. Furthermore, we explore the workflow of TinyML and analyze the identified deployment schemes, available tools and the scarcely available benchmarking tools. Finally, we discuss how reformable TinyML can impact a few selected industrial areas and discuss the challenges and future directions.
\end{abstract}



\maketitle

\section{Introduction}
The Internet of Things (IoT) paradigm is an emerging area that stimulates the growth of omnipresent and interconnected devices in various domains spanning from wearables to smart cities and infrastructure. Consequently, the growing market demand for IoT has resulted in it penetrating deep into our day-to-day lives. With this proliferation, vast quantities of heterogeneous devices produce tremendous amounts of equally heterogeneous data where concurrent Machine Learning (ML) and Deep Learning (DL) algorithms play a paramount role in processing this huge amount of data. For instance, Statista \cite{statista2025iot} assumed a total quantity of 11.7 Billion operational IoT devices (also referred to as things, intelligent devices/objects) by the end of 2020, whereas an estimation of 30.9 Billion was made for 2025. Comparing the time range, a substantial increase in operational smart devices is apparent. Quantifying the data generated, \cite{idc} forecasts an enormous 73.1 Zettabytes of IoT generated data by 2025. Previously, the severe resource and processing limitations of smart objects made the integration of Cloud Computers (CCs) a prospective approach as CCs possess considerable resources to empower resource-constrained devices in terms of processing, storage, and memory \cite{marotta2015managing}. However, the geographical displacement between things and CC inflicts delays that impede real-time, delay-critical applications and, as a result, affects the Quality of Service. 

However, today, the increasing scale and complexity of data have outdone CCs ability to keep pace. Moreover, increasing data breaches and CCs long-running reputation of acting as Single Points of Failure \cite{armbrust2010view} have led to a paradigm shift where data is processed in close proximity to the data generator, i.e., smart objects. Referred to as Edge Computing, the novel paradigm aims to address the perturbing challenges of CCs by providing storage and processing within the Radio Access Network \cite{abbas2017mobile}. Another definition by Shi \etal \cite{shi2016edge} states that Edge computing acts as an intermediary between data sources and CCs. From this, it can be deduced that interdependence between the two paradigms exists where traffic is sent to the Cloud regularly. However, the increasing amounts of data with cloud-dependent devices can potentially overburden cloud resources. A measure that could alleviate frequent access to the Cloud would be to integrate the processing mechanisms available in the Cloud (E.g., ML/DL) at the edge. By providing local cognition, intelligent things will be able to demonstrate advanced capabilities while disconnected, showing trace signs of intelligence; a sort of "Internet of Conscious Things" \cite{merenda2020edge}. Albeit, various technological barriers (E.g., infrastructure and network unavailability, cost per unit) hinder edge computing's broader yet potential impact. 

\begin{figure}
    \centering
    \includegraphics[width=0.75\textwidth]{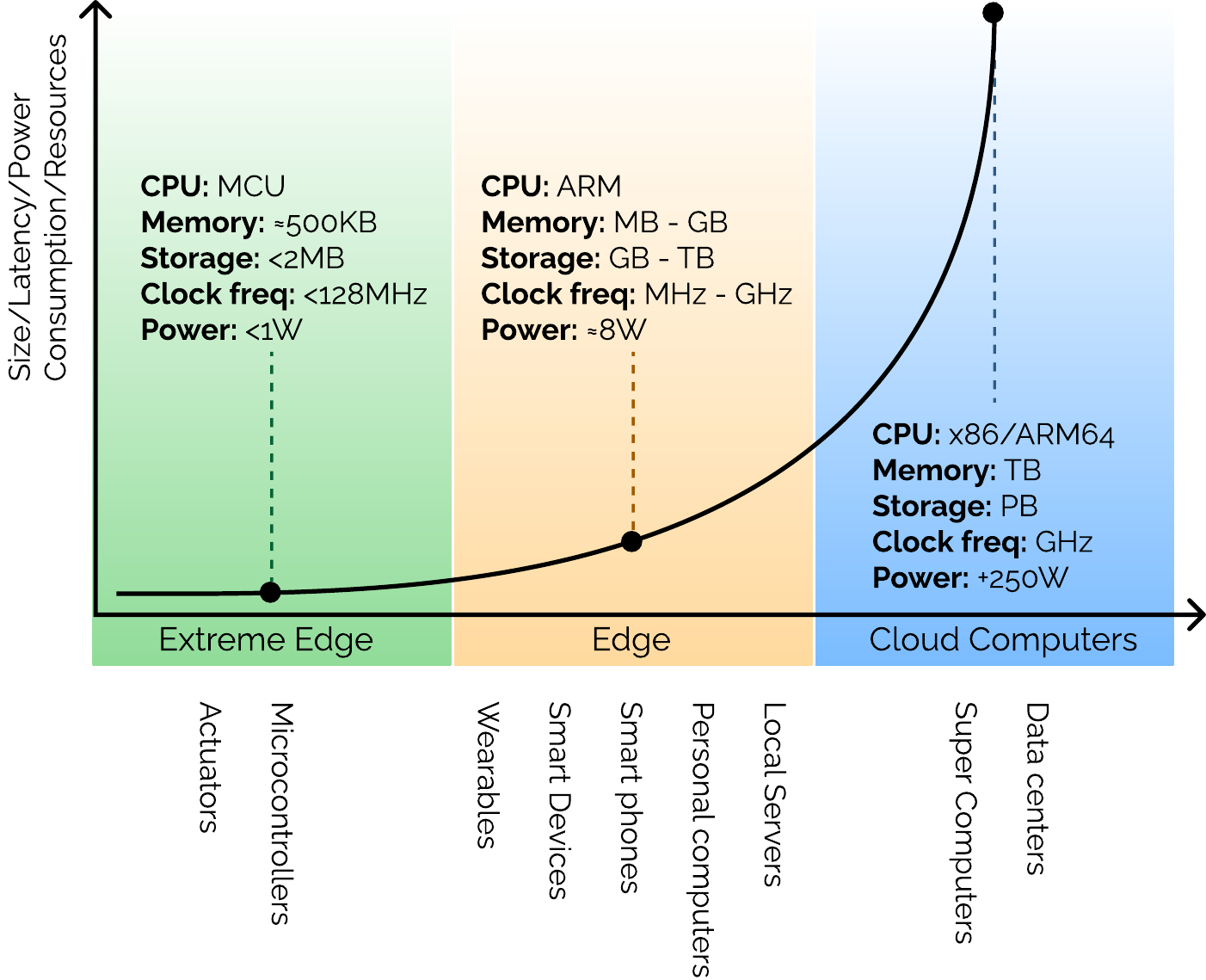}
    \caption{Comparison of paradigms}
    \label{fig:paradigms}
\end{figure}

In exploring alternatives to extend the IoT network, research has led to the exploitation of the MCU class of frugal devices for processing at the extreme-edge \cite{portilla2019extreme}. According to \cite{preden2015benefits}, pushing computation towards the absolute edge of the network, i.e., towards sensors and actuators, decreases latency while increasing autonomy. For a better perspective, refer to Figure \ref{fig:paradigms} that presents a stark comparison in resources, size, energy consumption and latency in each of the previously mentioned paradigms. Concretely, research efforts are now focusing on integrating TinyML as a remedial measure to alleviate some of the computational load accompanied by pure-edge systems that happen to burden CC resources.

TinyML is a nascent technology believed to be the forefront of hyper-digitization \cite{dutta2021tinyml}, where abundant MCUs will be exploited for executing ML at the extreme edge with the assistance of interfacing systems such as sensors. With $\approx$ 300 Billion existing MCUs \cite{icinsights2021}, harnessing the available idle resources for cognitive tasks could foreshorten the dependence of the cloud at the edge at a fraction of the cost and energy needed for deploying conventional edge appliances. For instance, Warden \etal \cite{warden2019tinyml} state that with most MCUs functioning in the Milli-Watt ranges, the industry has come to a common consensus that similar power consumption levels arbitrarily translates to an MCU having the ability to run on a coin cell battery for over a year. Paired with their inexpensiveness, TinyML allows a myriad of new always-on, inexpensive, battery-driven ML applications. It is worth indicating that although research in TinyML has seen explosive growth recently, implementations can be traced back to the advent of voice-based assistants. For instance, ML models used in Apples Siri and Googles Assistant exploit TinyML models popularly known as Keyword Spotting algorithms \cite{zhang2017hello} for assistant activation. Likewise, there have been a handful of solutions contradicting the stereotypical notion that ML/DL requires vast resources for processing. For example, in \cite{mohan2021tiny}, the authors miniaturized a CNN to a size of 138 Kilobytes (Kb) for face mask detection, a feat previously thought that only resource-rich devices could achieve.

Despite advancements in the domain, many existing solutions assume that TinyML only permits model inference provided by the scarcity of resources. Training is confined to resource-abundant devices such as Personal computers or CCs. However, solutions must consider the volatility and dynamicity of real-world data to produce effective ML solutions that are resilient to wear over time. We believe that for ML-driven extreme IoT to reach its full potential, it is crucial to experiment and invest in solutions that can be improved in time with minimal human intervention. 

Reformable TinyML is a countermeasure to the staticity as mentioned above. Static TinyML abides by the mantra "train-then-test" where once trained and deployed, updatability via network or through adaptivity are non-existent \cite{ravaglia2021tinyml}. Reformable TinyML, on the other hand, addresses issues in static TinyML and encompasses solutions that are capable of updating the models embedded to maximize performance in the environment they are placed in. 

Although there are a few surveys on TinyML, it is worth indicating that most focus on static TinyML \cite{dutta2021tinyml}, consider all TinyML implementations as a whole \cite{ray2021review, saha2022machine} \new{or survey a specialized domain in TinyML such as the available frameworks for development \cite{sanchez2020tinyml} or TinyML in healthcare \citep{tsoukas2021review}}.
They neglect to acutely discuss the potential of reformability and its broader implications in TinyML and other application domains. Furthermore, none of the existing surveys discusses the deployment methodologies for TinyML solutions. \new{Motivated by the lack of comprehensive reviews on the niche area of reformable TinyML, we aim to survey the new research bearing updatability/reformability of TinyML.} Our paper, for the most part, emphasizes reformable TinyML and discusses its industrial potential, challenges and future directions of it. In addition to that, we summarize the deployment methods and the minimal benchmarking frameworks available under the survey topic. In the current context, reformability is an extensive research area that, as of now, is seeing explosive growth. Because of this, it can be beneficial to have a taxonomy that describes reformable TinyML to ease the process of classifying and categorizing solutions. We present a newly composed taxonomy to fulfil this purpose, where we analyze the reforming methodologies, their integral components, and distinctive factors. In addition to the stated, we discuss the suitability of various approaches to provide reformability. \new{Table \ref{tab:related-publications} provides a succinct comparison of features between previous work and our paper to highlight its significance to the TinyML domain.}

\begin{table*}
\caption{\new{Comparison of related publications in terms of Reformable TinyML (Legend -- \cmark: Discussed, \xmark: Undiscussed, $\approx$: Partially Discussed)}} \label{tab:related-publications}
\begin{threeparttable}
\footnotesize
\scalebox{0.92} {
\begin{tabular}{ccccccc} 
\toprule
Publication & Year & Primary Focus & \makecell{Deployment \\ Methods} & Taxonomy & Benchmarks & \makecell{Next gen. \\ TinyML} \\
\midrule
Sanchez-Iborra \etal \cite{sanchez2020tinyml} & 2020 & TinyML Frameworks & \xmark & \xmark & \xmark & \xmark
\\

\midrule
Dutta \etal \cite{dutta2021tinyml} & 2021 & Static TinyML & \xmark & \xmark & \xmark  & \textbf{$\approx$}
\\

\midrule
Ray \cite{ray2021review} & 2021 & All systems & \xmark & \xmark & \cmark  & \xmark
\\

\midrule
Tsoukas \etal \cite{tsoukas2021review} & 2021 & TinyML in Healthcare & \xmark & \xmark & \xmark  & \xmark
\\

\midrule
Saha \etal \cite{saha2022machine} & 2022 & All systems & \xmark & \xmark & \cmark  & \xmark
\\

\midrule
Our Paper & 2022 & Reformable TinyML & \cmark & \cmark & \cmark & \cmark
\\

\bottomrule
\end{tabular}}
\end{threeparttable}
\end{table*}

The rest of the article is organized as follows.
Section \ref{overviewOfTML} provides an introductory overview of TinyML, which is essential to understand the "what's" and "how's" of TinyML. Section \ref{reformable-tml} introduces the concepts and motivations behind reformability in an IoT context and later introduces and delineates the proposed taxonomy. In Section \ref{sotaTML}, we systematically review the existing literature by the levels identified in the taxonomy. There, a summary is provided of the existing benchmarking frameworks for TinyML. Finally, in Sections \ref{potential} and \ref{challengesFutureTinyML}, we discuss the potential of Reformable TinyML in various applications and domains that currently exploit traditional TinyML, discuss the future directions for research, and finally analyze the role of next-generation computing in TinyML.

\section{Overview of TinyML} \label{overviewOfTML}

\subsection{Definition}

As the name suggests, Tiny Machine Learning brings cognitive capabilities to scarcely-resourced extreme IoT devices such as Microcontroller units.
The term "TinyML" was coined by Warden \etal for ML applications that use miniaturized models for extremely energy-efficient inference \cite{tinyml_webpage,reddi2021widening}. Further, the lack of resource-rich Operating Systems (OS) (E.g. Linux, Windows and macOS) on MCUs resulted in Doyu \etal \cite{doyu2021tinymlaas} extending the definition by Warden \etal as \textit{the intersection between ML and constrained-IoT devices without a resource-rich OS}. \new{A plethora of devices falls under the category of MCUs, including but not limited to the Arduino UNO, ESP8266, and ATmega328. Several other TinyML-compatible boards can be found in the research attempts reviewed below or on the Tensorflow Lite website \footnote{https://www.tensorflow.org/lite/microcontrollers}}

Concretely, the roots of TinyML stem from the advent of EdgeML, which mitigates the infrastructural needs to persistently transfer/receive data of conventional IoT. Similarly, TinyML is not considered a substitute for the Fog or Cloud paradigms; rather a supplementary component that works in tandem with the said paradigms.
 
\subsection{Workflow}

Embedded systems are abundantly heterogeneous frugal devices that range from a few Kilobytes to a few hundred Kilobytes in memory, $<$2 Megabytes of flash storage, and a few Milliwatts ($<3$mW) of energy usage. Traditional mobile devices, which are gaining in computational capabilities and energy efficiencies, still draw hundreds of Milliwatts to a few Watts for operation. These power usages are significantly higher than what MCUs consume \cite{banbury2021micronets, fedorov2019sparse}. 

\begin{figure*}
    \centering
    \includegraphics[width=\textwidth]{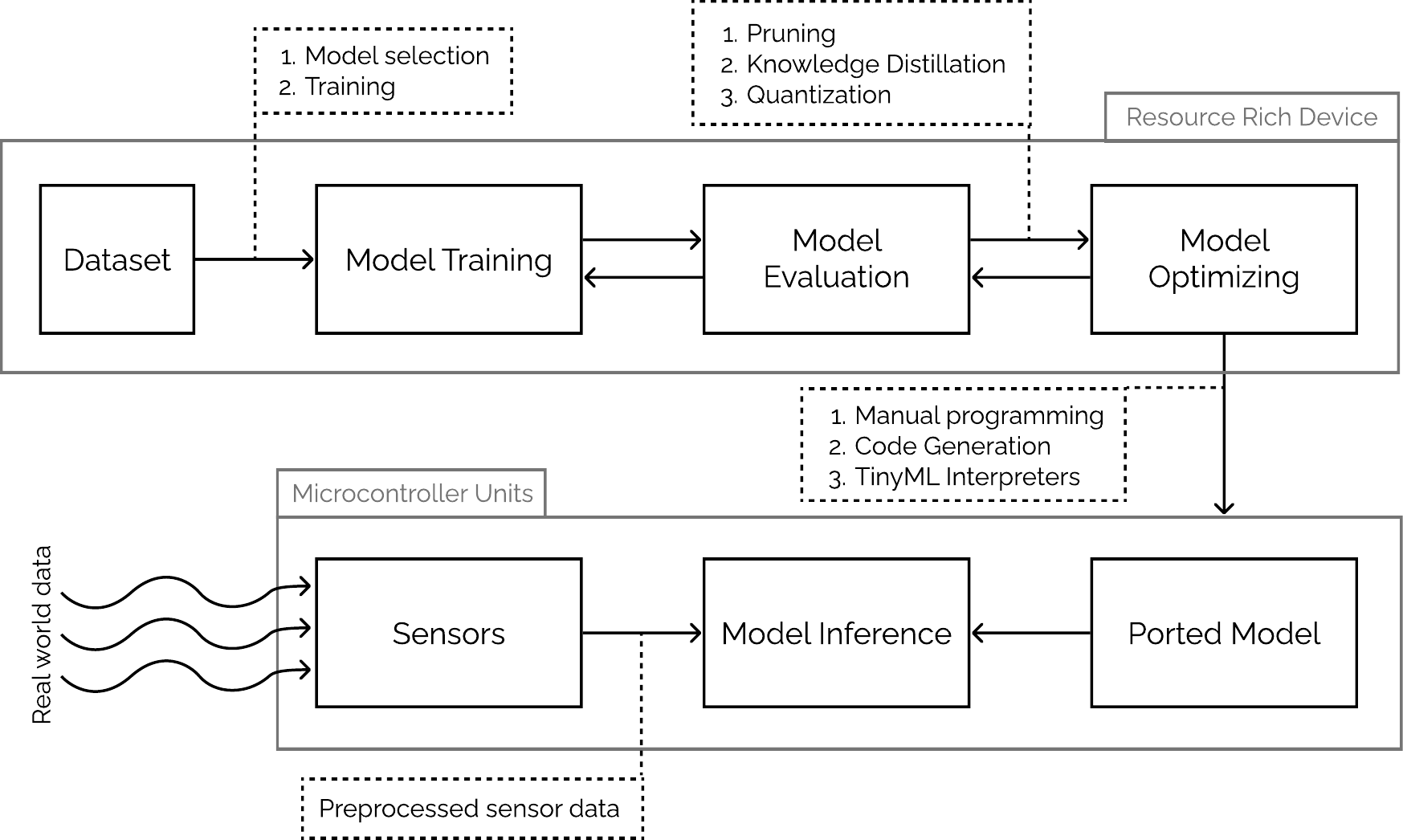}
    \caption{Generic TinyML Pipeline}
    \label{fig:workflowTML}
\end{figure*}

The conventional workflow for deploying TinyML can be observed in Figure \ref{fig:workflowTML}. The process is divided into three distinct phases: training, optimizing and deploying. However, the architecture can be abstracted into traditional ML and TinyML components. The former encapsulates data collection, algorithm selection, model training and optimizing phases which are de-facto conventional in traditional ML approaches. The latter specializes in TinyML and consists of the model porting and deploying phases.

The preliminary step in developing a TinyML model is to select a relevant algorithm and collect the data for it. Precompiled data repositories or sensor data gathered on the fly can be leveraged to fulfil this purpose. For instance, the authors of \cite{zhang2017hello} used the Google Speech Commands repository to train and evaluate a Key-Word Spotting algorithm. Alternatively, \cite{prado2021robustifying} utilized a sensor-composed dataset for training a Computer Vision Algorithm for self-driving micro-cars. Once the necessities are gathered, the model training process can be conducted in a resource-rich device (e.g., server or personal computer). Like traditional ML, contemporary ML frameworks such as TensorFlow, Pytorch, Scikit-Learn can be employed for training.

Once the model training is completed, the ensued model needs to be optimized. Model optimization can be done through Pruning, Knowledge Distillation, Quantization and Encoding techniques to attain the desired requirements \cite{dutta2021tinyml}.
\textbf{Pruning} is an iterative process of systematically removing parameters from a Neural Network (NN) \cite{blalock2020pruning}. 
\textbf{Knowledge Distillation (KD)} is teaching a small yet well-optimized model (student model) to mimic the activations of a pre-trained model (teacher model) \cite{park2019kd}. Unlike Transfer Learning, where the model architecture remains untouched, KD maps the generalizations into a lighter model.
\textbf{Quantization} refers to the reduction of numerical precision of a network. For instance, the weights and activations are often represented in floating points (32/64 bit). By representing the floating-point values as integers, a model can be classified as quantized \cite{cai2020zeroq}. However, close attention needs to be paid as quantizing a network from single precision to low precision can degrade performance.

Porting and deploying are the final steps in a TinyML solution. A model needs to be ported to an MCU understandable language before deploying. Typically, C/C++ is the prevalent language for such applications. For this case, TinyML interpreters (E.g., Tensorflow Lite Micro \cite{david2021tensorflow}) are used to convert a model built using an available programming language to a frozen graph ideally represented as a C array \cite{ren2021tinyol, disabato2020ondevice}. 

Finally, the ported model is embedded into an MCU, where inference is performed on sensor data with the help of the underlying ML mechanism. A broader survey on the available porters and TinyML compilers are discussed by Sanchez-Iborra \etal in \cite{sanchez2020tinyml}. The deploying methodologies are discussed in further detail in  Section \ref{deploying-methods}.

\subsection{Deployment Methods} \label{deploying-methods}
The widespread adoption of embedded solutions advocates continuous research in low-resource technologies that can be brought under the general criteria of TinyML. A critical component of TinyML solutions is the embedded inference mechanisms where research is generous in numbers. It should also be noted that models generated by popular ML frameworks such as Tensorflow, Pytorch, and SKLearn are unsuited for TinyML in its raw form due to their memory requirements. As a result, multiple runtimes have been proposed to support ML/DL on MCUs. However, only three distinct methods have been identified for deploying TinyML solutions. Namely, Manual Programming, Code Generators, and ML Interpreters \cite{banbury2020benchmarking}.

\textbf{Manual Programming} yields the best results primarily due to the possibility to tinker all aspects for optimizing purposes selectively. Despite the potential to produce research, sharing knowledge is often limited. This is due to the optimizations made being obscured and confined to a selected party. Moreover, the heterogeneity of MCUs hinders the capacity to unify these specialized Embedded ML approaches. From a functional standpoint, TinyML solutions will benefit in terms of performance from this approach at the cost of reproducibility and time, making such implementations less than ideal for non-domain experts (E.g., Hobbyists). However, from a reformable perspective, the lack of toolsets to update models on the fly make manual programming approaches the only alternative.

\textbf{Code Generation} tools are favored in terms of convenience as an optimum solution can be reached, depriving the need to program it manually. Many of the code generation tools such as EdgeImpulse \cite{edgeimpulse}, Imagimob \cite{imagimob} exploit AutoML to provide the necessary tools as Software-as-a-Service, whereas \cite{cartesiam} acts as a search engine for third party TinyML libraries. A drawback of Code Generation approaches is that the fragmented market and vendors have proprietary toolsets and compilers that make interoperability and portability a challenge \cite{banbury2021micronets}. The available code generation tools are summarized in Table \ref{tab:code_generation}.

\textbf{Interpreters}, in contrast, offers superior portability as the structure is similar regardless of the device. ML/DL models can be easily ported as a model's architecture is not coupled to the framework. Additionally, the models can be individually optimized and fine-tuned to suit the needs and devices. Despite the convenience, this scheme comes with a small overhead in performance and memory usage. Further, the separation between model and system-level processes allows for greater generalizability, thus making it easier to introduce benchmarks.

As interpreters can make embedded machine learning accessible to all, well-established technological companies are taking open source initiatives to unify TinyML. For instance, Google lead with TensorFlow Lite Micro (TFLM) \cite{tflite}, and Microsoft with Embedded Learning Library (ELL) \cite{msell}. In the line of available frameworks, a detailed survey conducted by Sanchez-Iborra \etal \cite{sanchez2020tinyml} compares the many TinyML frameworks available.

\begin{table*}
\caption{Comparison of Code generation tools for TinyML} \label{tab:code_generation}
\centering

\small
\scalebox{0.92} {
\begin{tabular}{cccccc} 
\toprule
Tool & \makecell{Specified \\ Supported \\ Architectures} & Language & Compiler & Runtime & Specialization\\
\midrule
EdgeImpulse\cite{edgeimpulse} & \makecell{Cortex-M7 \\ Cortex-M3 \\ Cortex-M4 \\ Cortex-M4F \\ Cortex-M33 \\ Cortex-M0+} & C++ & EON & TFLM & \begin{minipage}[c]{0.4\textwidth}
    \begin{enumerate}
        \item Freely available.
        \item Prebuilt models cannot be imported but the architectures can.
        \item Converts the NN to source code for 20-30\% reduction in memory usage \& eliminates the need for the TFLM interpreter.
    \end{enumerate}
  \end{minipage} \\
\midrule
Imagimob\cite{imagimob} & \makecell{Cortex-M0 \\ Cortex-M3 \\ Cortex-M4} & C &  N/A & N/A & \begin{minipage}[c]{0.4\textwidth}
    \begin{enumerate}
        \item Subscription based
        \item Relies on AutoML based C code generator named Imaginet.
        \item Supports post-training quantization for Long-Short Term Memory layers.
        \item Eliminates the dependence on runtimes as code is statically compiled.
        \item Utilize "imunits" for optimizations.
    \end{enumerate}
  \end{minipage} \\
\midrule
NanoEdge AI\cite{cartesiam} & All Cortex-M & C &  N/A & N/A & \begin{minipage}[c]{0.4\textwidth}
    \begin{enumerate}
        \item Search engine for static TinyML models
        \item Ability to limit memory allocated for the model
        \item Not limited to NN implementations
    \end{enumerate}
  \end{minipage} \\
\bottomrule
\end{tabular}}
\end{table*}

\subsection{Characteristics of Embedded Systems}

Embedded systems (used interchangeably with MCUs), in general, have a few distinct features that are immediately noticeable. These features make them a valid candidate for various intelligent edge applications in hindsight. The advantages and disadvantages of such systems are elaborated below.

\subsubsection{\textbf{Energy Efficiency}} \label{energyEfficiency}

As mentioned previously, the immediately apparent advantage of MCUs is their superior energy efficiency. An MCU will draw significantly lower energy even at the maximum workload compared to conventional edge devices. The low power usage provides MCUs with the ability to power themselves using miniature batteries or energy harvesting methods \cite{garg2017energy} (e.g., solar, vibrations, thermal) \cite{sanchez2020tinyml}. Additionally, the scarce energy signature correlates to the devices' small dimensions, making MCU's a viable option for placing them in remote locations without a constant connection to a power grid -- giving them the name of Peel-and-Stick sensors \cite{lalau2017peel}. Moreover, most devices can foster an even lower energy signature by eliminating the need for Radio Access Technologies (RATs) to communicate with external entities.

\subsubsection{\textbf{Placement}}

The miniaturization of System on Chips due to advancements in high-throughput, energy-efficient architectures makes it possible to fit sufficiently resourced processors on devices where placement options are limitless. For instance, \cite{wu2018implantable} designed a sensor that can be implanted under the skin that is powered by subcutaneous energy harvesting. The above is a testament to such systems' compactness and energy efficiency.

MCUs are abundantly available and easily accessible, which is a direct result of the low cost ($\approx1$\$) \cite{icinsights2021} of these pervasive devices. This cost-effectiveness encourages use cases such as low latency analysis and modeling of sensor signals from sensors in the manufacturing industry, agriculture, e-health, or entertainment\cite{sanchez2020tinyml}. As a result, this nurtures the notion among industries to replace electro-mechanical or analog circuitry with software-defined alternatives \cite{david2021tensorflow}. Moreover, re-programmability and ubiquity bring endless opportunities to provide highly dynamic intelligent systems.

\subsubsection{\textbf{Heterogeneity}} \label{heterogeniety}
The differences within embedded devices aid the widespread adoption among various industries. As a result, a particular disadvantage of embedded devices is that the market is heavily fragmented in terms of manufacturer and use case. As a byproduct, countless devices with vastly heterogeneous resource capacities are made available for choosing, thus making benchmarking a tedious hurdle to overcome.

\subsection{Characteristics of TinyML infused MCUs}

Previously, we discussed the distinctive features of embedded systems. Embedded systems are great candidates for alternative processing at the edge. Coupling MCUs with cognition can allow for new application use cases. Further, we can notice previously unseen characteristics at the extreme edge, which will be discussed here.

\subsubsection{\textbf{Responsiveness}} 
Most networks formed using low-powered embedded devices are segregated into Wireless Sensor Networks (WSN) or Low power and Lossy Networks (LLNs). Intermediate connections in LLNs are characterized by high packet loss rates, low bandwidth, and instability \cite{sankar2017lln} making such networks impractical for delay-critical and connection-critical applications. Similarly, WSNs rely on a centralized primary node, which causes delays in processing the data based on its capabilities. A proven alternative is to process the sensor data locally, which benefits responsiveness. Nevertheless, improved responsiveness will play a critical role in the future. For instance, offloading a task in Augmented Reality (AR) glasses to the cloud, edge, or a neighboring device can cause delays that could cause an unpleasant user experience. The necessity for always-on, battery-driven processing in AR proves TinyML to be a suitable candidate \cite{banbury2020benchmarking}.

\subsubsection{\textbf{Privacy}} 
Ensuring data privacy and security is an inherent advantage of processing data on-premise. Data can be sensitive; therefore, localizing the processing eliminates the need to transmit raw data to or from a central entity. Consequently, this reduces the chances of privacy and security breaches.

Further,  duty cycling \cite{zhang2017dutycycle}, i.e., scheduling the power on/off states of a RAT for radio communication, can be employed for additional security and improved longevity in devices capable of wireless communication. The ability to toggle the state of the RAT aligns well with concepts such as Collaborative Machine Learning (discussed further in Section \ref{distributedTML}), which focuses on improving a global model as a collective.

\new{Before advancing, it must be noted that even though many existing implementations surveyed in this article focus on research on the advanced topic of reformable TinyML, several resources have been authored or created in attempts to make TinyML beginner friendly. The books TinyML Cookbook \citep{gianmarco2022cookbook} and TinyML \citep{warden2019tinyml} provide a concrete foundation for TinyML by discussing the history and practicality and, after that, gradually transition to real-world applications and implementations which are publicly available. Apart from books, organizations such as TensorFlow and EdgeImpulse advocating TinyML have working projects and examples with their respective documentation available for reading \cite{tensorflowex, edgeimpulseprojects}. These cited resources are only in the domain of static TinyML. The existing reformable TinyML attempts primarily exist in research with marginal resources that beginners or intermediates can use/follow to replicate. However, some researchers have followed the trend of making their work open source on git repositories. Work on reformability that has open repositories can be found in Table \ref{tab:sota-performance}.
}

\section{Introducing Reformable TinyML} \label{reformable-tml}

\new{In this section, we discuss Reformable TinyML and present a novel taxonomy for classifying reformable TinyML solutions. We will start with some introductory information on Local and Remote updates, the idea of concept drift, as well as the article selection and filtering criteria used in this paper.}

\subsection{Overview}

Amidst the dynamicity of an environment, IoT encounters diverse data. Given the current context of ML-driven IoT, staying up-to-date benefits stakeholders across the spectrum. Customarily, updates can be segregated into Local and \new{Remote/Over-the-Air (OTA)} updates \cite{halder2020ota}. Many deployed IoT applications resort to OTA updates mainly due to the lack of resources for local updates, meaning that so far, only algorithms running inference on resource-rich devices would have reaped the benefits of local updates \new{such as Online or Lifelong learning}. Nevertheless, the advancements made in the field are endowing resource-constrained IoT with the ability to learn on the fly. Similarly, the research in TinyML has led to breakthroughs in Reformable TinyML, i.e., TinyML solutions that can improve themselves via local or OTA updates.

Evidently, most of the implementations assume that TinyML is static, i.e., only runs inference. However, TinyML, in particular, can benefit from reformability. Provided that MCUs can be placed in hard-to-reach locations (E.g., implanted sensors, machinery) and notably, an MCUs peel-and-stick nature, make reformability a potential candidate. Learning from new data points or wirelessly updating models can prevent iterative extracting, training, and deploying seen in typical batch-learning TinyML solutions. Furthermore, the long deployment periods achieved via batteries or energy harvesting in TinyML translate to online learning models being exposed to higher quantities of data used for training. Thus, the more significant potential to yield better results. Despite integrating reformability in a resource-constrained setting like an MCU being daunting, research in the area has seen explosive growth recently.

It is worth noting that the foundation of reformable ML, in general, originates from the need to combat Model Drift -- the inevitable degradation of a model's performance due to the ever-changing nature of data. Model Drift/Decay predominantly exists in highly volatile environments where data is prone to rapid or random fluctuations. This phenomenon is commonly referred to as \textbf{Concept Drift}.

Gama \etal define Concept Drift as the change in the relation between inputs and target variables underlying data distribution \cite{gama2014adaptation}. Some use the term Dataset shift \cite{quinonero2009datasetshift} to define the observations as to when the distribution between the input and output data are inconsistent between the training and deployment phases. A concrete example of such would be an online clothing store. A recommendation system that recommends items based on a user's interests in the summer will yield inaccurate/poor performance in the winter. 

\textbf{How were articles selected for the survey?} \new{In order to conclude our preliminary search for relevant articles, Google Scholar was used as the primary database. Apart from Google Scholar, other databases and tools such as Semantic Scholar, Papers with Code, Connected Papers, IEEE Xplore, ACM Digital Library, and Elsevier were utilized. Compiling the final selection of articles required a rigorous search process done by combining several keywords, including but not limited to the phrases TinyML, Incremental learning, Over the Air, updates, Concept drift, Online learning, Federated learning, and Collaborative ML. Then we reviewed the titles, abstracts, and conclusions to ensure the article's relevance to the reformable scope. Next, we went through the references of the selected articles in an iterative process to select more relevant articles. The following criteria were considered for the final shortlisting.}
\begin{itemize}
    \item \new{Is the article in the domain of IoT?}
    \item \new{Does the article focus on running ML/DL on extremely energy-efficient devices such as MCUs?}
    \item \new{Does the article research on incremental/online learning, OTA, Collaborative TinML solutions?}
\end{itemize}
\subsection{Taxonomy of Reformable TinyML} 
\begin{figure*}
    \centering
    \includegraphics[width=\textwidth]{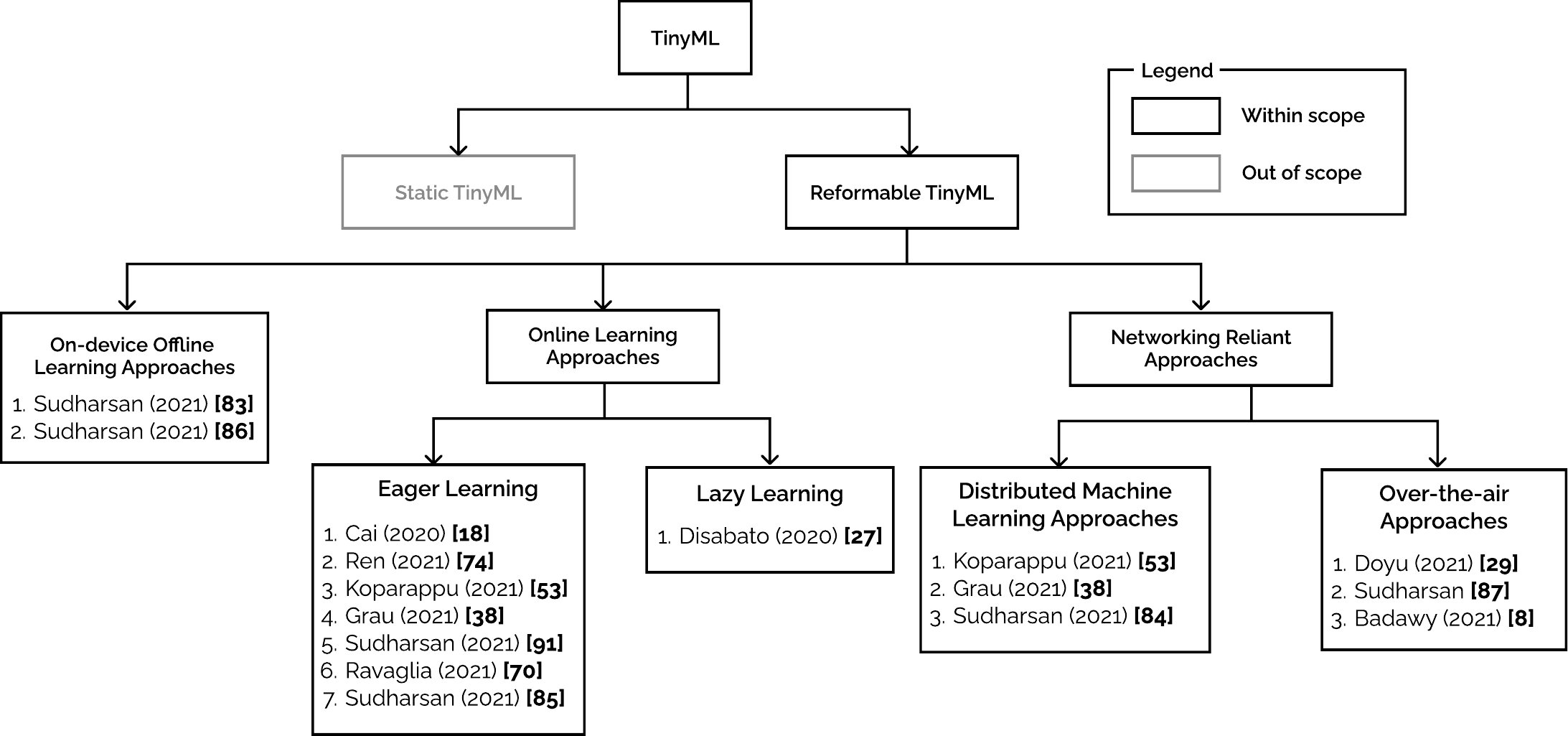}
    \caption{Taxonomy of Reformable TinyML}
    \label{fig:Taxonomy}
\end{figure*}

There has been explosive growth in research encapsulating Embedded Machine Learning and extreme IoT. However, existing research on model adaptivity is scarce and heavily fragmented in methods and techniques. 
To elucidate, this section presents a taxonomy of reformable TinyML solutions (See Figure \ref{fig:Taxonomy}). The primary focus is to acutely perceive the distinctive factors of each of the hierarchical levels. Research on reformable TinyML can be segregated into Online Learning, and Network Reliant approaches. Each of the approaches in the secondary level can be divided further: Online Learning into Eager and Lazy Learning techniques and Network reliant approaches to Distributed Machine Learning Approaches and Over-the-air approaches.

\new{\textbf{On-device Offline Learning approaches}: Conventionally, offline learning approaches are more prevalent in ML. Also known as batch learning, these techniques refer to models that are trained by utilizing a pre-compiling dataset, i.e., a static dataset \cite{bisong2019batch}. A distinguishing factor between online and offline learning approaches is that the model's parameters are only updated once the entire batch has been fed through the model. In contrast, online learning learns after each training data instance/data point. Although offline learning hardly classifies as reformability, there have been successful attempts to train an offline model within the premises of embedded devices resource constraints which is an impressive feat given the context. However, the overall feasibility of on-device offline learning approaches is yet to be questioned due to the resource overhead.} 

\textbf{Online Learning Approaches}: This approach is commonly referred to as Incremental Learning; this approach successively updates a model as data often arrives sequentially. To elaborate, when a new data point arrives, the model is updated in place to produce the "most" up-to-date model at that given instance \cite{chen2018lifelong}. Put simply, given the context, the model assumes to be the best version of itself. These algorithms, unlike Lifelong Learning \cite{parisi2019continual} that mimic a human's physical capability to learn from multiple tasks continually, then store and distribute knowledge, only adapt to a singular task (E.g., Classification of dogs, flowers). Online learning benefits in use-cases where accommodating a dataset in the primary and secondary memories is infeasible (E.g., IoT, Wearables, Sensor units). 
In such scenarios, to preserve memory and processing, the real-time data points have a limited number (often one) of processing passes \cite{gama2014adaptation}.

\begin{enumerate}
    \item \textbf{Lazy Learning} is a learning method, which directly stores the data point and defers generalizations (processing) till it is queried later \cite{wettschereck1997lazylearn}. By definition, Lazy Learning algorithms require a knowledge base to store data samples; hence, scaling poorly as the number of data samples grows. Lazy Learning can be less than ideal in an MCU setting as the requirement of a knowledge base on-premise becomes increasingly difficult to satisfy with growing samples. The most notable example for such an algorithm is the $k$NN classifier.
    
    \item Contrasting Lazy Learning, \textbf{Eager Learning} constructs a generalized, input independent concept description (typically represented by NNs, Decision trees, Naive Bayes) based on the training data \cite{aha2013eagerlearning}. A majority of existing TinyML solutions rely on NNs for processing sensory input. Thus, using this technique is a viable option to revamp existing systems. However, in comparison to Lazy Learning, Eager Learning approaches need higher processing power and time for training but drastically less time for predicting \cite{bhavsar2012comparative}. Typically, Backpropagation accounts for most of the elapsed training time \cite{bhavsar2012comparative}. Increased processing capabilities are vital for speedy training, thus making the increased training time another potential roadblock for complex NNs in TinyML. However, with no requirement for a knowledge base, these techniques can learn from a limitless quantity of data.
\end{enumerate}

\textbf{Network Reliant approaches}: This classification of approaches relies on external entities to update TinyML models, necessitating a networking component to communicate with the relevant resource-rich entity. Frequently, these external entities leveraged predominantly consists of Cloud computers or Fog nodes \cite{bonomi2012fog}. Network reliant approaches can be divided into Distributed TinyML and Over-the-air approaches.

\begin{enumerate}
    \item \textbf{Distributed ML} is another nascent research area that has seen growth influenced by its privacy-preserving nature. This approach focuses on collaboratively training models on decentralized data, i.e., training closer to the data source. Popularly, there are two variants of Distributed learning suitable for edge applications. They are Federated Learning (FL) and Gossip Learning (GL) \cite{hegedHus2019gossip}. FL was first introduced in the seminal paper by McMahan \etal \cite{mcmahan2017communication}. Here, the authors advocate the distribution of training at the edge where local updates are aggregated to a \textit{central} server. GL, on the other hand, has no central aggregator to collect local changes, i.e., Decentralized. As the term "Gossip" implies, the models in GL are sent to random nodes in the network for model convergence. In the context of TinyML, FL is the better candidate out of the two approaches. In GL, model convergence occurs among edge nodes selected randomly, meaning that compared to FL, the communication overhead is significantly higher in GL. Furthermore, randomness in selecting the next node could lead to disparities in energy levels between regularly and rarely selected nodes affecting longevity. 
    
    \item An alternative to Distributed ML mechanisms is the using \textbf{Over-The-Air (OTA)} updates for reforming models. Unlike Distributed ML, OTA updates provide the ability to resolve bugs and security vulnerabilities identified post-deployment or even support completely different functionality \cite{arakadakis2021firmware}. In MCUs, the firmware has to be "flashed" into a programmable memory, i.e., loading the update's contents into the secondary, involatile memory.
\end{enumerate}

It is noteworthy that even though Distributed TinyML can work similar to Online Learning, i.e., storing and updating the model in the primary memory, they can also adopt OTA approaches to embed the new model in the secondary storage. Given this, most MCUs consist of either Flash storage or Electrically Erasable Programmable Read-only Memory (EEPROM). Both of which have a limited number of write cycles known as "endurance," based on the chipset \cite{aras2020microvault}. Given the context, constant writes to the secondary storage in MCUs will eventually result in the memory deterioration to an unusable state over time. As robust as it is for TinyML, it is essential to consider the drawbacks of the storage mechanisms when developing systems utilizing the approaches discussed.

\section{State of the Art in Reformable TinyML} \label{sotaTML}

The dawn of TinyML has exposed new research avenues in ML-driven IoT. With the growing interest and potential, pertinent research has led to recent research accomplishing state-of-the-art status in adaptive TinyML. 
This section distills the advancements in the topic with a comprehensive analysis of the existing benchmarking suites. \new{The Tables \ref{tab:sota} and \ref{tab:sota-performance} summarizes the following work in succinct in terms of architecture, use-cases, specialization} 

\new{\subsection{On-device Offline Learning} \label{offlineLearning}
It is well known that offline learning, in any circumstance, requires resources for training and especially to host a dataset. As impossible as trying to replicate offline learning on an MCU appears, Sudarshan \etal attempted to contradict the above statement with Edge2Train \citep{sudharsan2020edge2train}. Edge2Train introduces an offline learning framework based on Support Vector Machines (SVMs). The architecture of Edge2Train fuses four blocks where each harmoniously contributes to the improvement of the model. These blocks are 1) The I/O block, which acquires and pre-processes sensor data into actuation commands, i.e., signals that allow peripherals to interact in the real world and feed it to the model. 2) A feedback block that monitors each real-world actuation made by the model in the background gives feedback on whether the actuation was satisfactory based on its observations. The output of this block is considered the ground truth. 3) The Data-stitching block is responsible for creating a local dataset. It uses the model's input and output arrays for an actuation flagged as unsatisfactory by the feedback block alongside the ground truth 4) Finally, the Train-on-demand block trains and evaluates a new model with the dataset compiled when the existing model's performance diminishes. The new model is then replaced with the pre-existing model. Two i7 CPUs and five MCUs with different specifications were selected to evaluate the models. The authors finally conducted a thorough evaluation where the same SVM parameters, hyper-parameters and datasets (Iris and MNIST datasets) were used across the CPUs and MCUs. Performance-wise, the highest classification accuracy for the Iris dataset on the CPUs and MCUs was 96.67\% and 90.0\% , respectively. For MNIST, the highest accuracies for the CPUs and MCUs were 92.85\% and 91.66\% , respectively.}

\new{The authors of Edge2Train have also included an evaluation of energy consumption (in Joules) between CPUs and MCUs. For the MCUs, they multiplied the current rating (in Amperes) of the MCUs with the voltage (in Volts) and the time elapsed by the task (in seconds). Equation \ref{energymeasure} formulates the equation used by Sudarshan \etal to estimate energy consumption. For the CPUs, the powerstat tool on Linux and htop process viewer on Windows were used.}
\begin{equation}
\label{energymeasure}
\begin{split}
Energy(J) = Power(W) \times Time(s) \\
Power(W) = Current(A) \times Voltage(V)\\
Energy = (Current(A) \times Voltage(V)) \times Time(s)
\end{split}
\end{equation}

\new{Results show that at its most efficient, the MCUs consumed $2\times$ more energy to train on the MNIST compared to the Iris dataset. Adding to this, the MCU consumed 3.4$\times$ more energy to run inference on MNIST over the Iris dataset. Finally, the authors conclude that the MCU consumes $\approx20\times$ and $\approx350\times$ less energy than the CPU for a unit inference.}

\new{Like Edge2Train, Sudharsan \etal formulated MCU-ML \citep{sudharsan2021mlmcu} that seemingly improves from Edge2Train by providing developers with a framework that allows them to train binary classifiers or multi-class classifiers on-device. An algorithm dubbed Optimised Stochastic Gradient Descent (Opt-SGD) enables the training of binary classifiers, which inherits beneficial aspects of both Gradient Descent and Stochastic Gradient Descent. For multi-class classifiers, Optimized One-VersusOne (Opt-OVO) is introduced. Opt-OVO allows training by removing base classifiers that do not contribute to the final results. Furthermore, Opt-SGD is core to Opt-OVO as it pushes training efficiency even further by decomposing the multi-class problem into several binary class problems. The authors conduct several evaluations for the two algorithms using four datasets and four MCUs. Results show that on-device training using Opt-SDG and Opt-OVO is feasible, even on low resourced devices with decent performance.}

\subsection{Online Learning} \label{adaptiveLearning}
As previously mentioned, Online/Incremental Learning is ideal in settings where memory and storage are restricted. As per the existing literature, Online Learning approaches can be divided into Eager, and Lazy Learning approaches.

\subsubsection{\textbf{Lazy Learning Approaches}}

To recall, Lazy Learning techniques store the training data till testing data is exposed to the system. Then, these algorithms make estimations for classification based on the samples in the knowledge base. In the context of TinyML, employing a knowledge base for Lazy Learning can be strenuous. Despite this, Disabato \etal produced the first and only Online Lazy Learning solution in TinyML.

They \cite{disabato2020ondevice} deployed the first Online Learning algorithm in the context of TinyML on 2 off the shelf systems, i.e., a Raspberry (RB) Pi 3B+ and an STM32F7 MCU. The authors devised a lightweight incremental mechanism that utilizes a deep transfer learning-based Feature Extractor and a $k$NN classifier to address two binary classification problems. Key-Word Identification and Image classification use cases were considered for this specific implementation. Online learning is achieved by following a "test then train" approach where supervised data is provided to the $k$NN rarely, post-classification. Furthermore, a non-trainable DL-based Feature Extractor based on Task Dropping was introduced. $k$NN classifier as the "learner" was chosen under the basis that it is a non-parametric classifier, which needs little training. The experiment was conducted in Raspberry Pi 3B+ and STM32F7 devices. Compared to a pre-trained Support Vector Machine (SVM), the $k$NN provided lesser accuracies (5-10\% less) on the RB Pi 3B+. The accuracy metric was disregarded for the STM32F7 as the replicated architecture does not affect accuracies. However, the authors evaluated computational times on this device to demonstrate it's feasibility. Here, the authors noted that the processing time associated by the feature extractor surpasses the $k$NN classifier.

\subsubsection{\textbf{Eager Learning Approaches}} \label{eagerLearning}

Like Lazy Learning, Eager Learning post-deployment in TinyML is faced with its challenges. In general, Eager Learning inference times are a fraction of the time to train the model. The same applies to TinyML, except limitations in processing, storage, and memory need to be accounted for, and special attention must be given. Regardless, research on Eager Learning Online TinyML has shown explosive growth in recent years.

Cai \etal \cite{cai2020tinytl} observed that the training memory overhead is influenced primarily by the activations. Considering this, they introduced Tiny Transfer Learning (TinyTL), where the weights were frozen while the bias' incrementally learned, thus mitigating the need for the intermediate activations. Additionally, they composed a memory-efficient bias module named Lite Residual Module for the incremental learning task. Results show that the TinyTL reduces memory usage for training significantly. Similar to Transfer Learning, Ren \etal \cite{ren2021tinyol} devised a layer (named TinyOL) that can be trained on streaming data. The layer achieves adaptivity by running in the RAM.

Furthermore, TinyOL can be inserted into an existing network where the newly attached layer will function as the network's output layer. Ren \etal assessed the system in a supervised (Anomaly Classification) and unsupervised (Anomaly Detection) setting for evaluation. However, the results are in favor of the model trained offline. 
Sudarshan \etal in \cite{sudharsan2021trainplusplus} produced Train++, a custom resource-friendly binary classifying algorithm that has onboard training enabled. It should be noted that, unlike many of the previous Eager Learning approaches, Train++ is a non-NN implementation. As evaluations show for five different MCUs, a 5.15 - 7.3\% increase in accuracy and $\approx$10 - 226-second reduction in training time can be seen across the board for the seven datasets used. \new{Notably, the authors have compared the estimated energy consumption between Edge2Train, \citep{sudharsan2020edge2train}(Section \ref{offlineLearning}) and Train++ using Equation \ref{energymeasure}. Results show that Train++ consumed $\approx34000 - 65000\times$ less energy for training and $\approx34 - 66\times$ for single inferences depending on the use-case.}
Ravaglia \etal \cite{ravaglia2021tinyml} on the other hand, developed the first hardware/software integrated platform for Continual Learning at the extreme edge. They achieved this with Quantized Latent Replay-based Continual Learning (QLR-CL) -- a low-bitwidth quantization process based on Latent Replay rehearsing to reduce the memory overhead for Continual Learning. Experimentation was conducted on a ten-core, 32-Bit Parallel Ultra-Low-Power (PULP) device\new{, dubbed VEGA,} with significantly higher resources than MCUs with the STM32 architecture. Demonstrations show that by reducing the precision of the model between 8-Bit and 7-Bit, an accuracy drop of 0.26\% and 5\% can be observed. They finally conclude that an $\approx65\times$ performance gain can be seen compared to MCUs with STM32 architectures. \new{To evaluate energy consumption, the authors study two use cases. The first is a single mini-batch of the Core50 dataset, i.e., a dataset and benchmark specifically crafted for testing and assessing continual learning in object recognition tasks. The second use case is a simplified study of the demonstration done by \citep{pellegrini2020latent}, which displays the ability of a model to adapt to new classes and improve its performance on existing ones. Furthermore, the authors also run and compare the results of the research implementation on another energy-efficient MCU: NUCLEO-64. Only the first experiment's last layer (layer 27) was retrained. It must be noted that these tests only provide a 3300 mAh battery as an energy source. The final results show that VEGA can perform over 1080 retraining events per hour, giving the device a lifetime of $\approx175$ hours. The authors also noticed that training larger segments of the network, the training time eventually results in a learning event rate of about 10 per hour, which can extend the lifetime to 200-1000 hours. When the final layer of the NUCLEO-64's model is retrained, the maximum retraining event rate is set to 750, reaching a lifetime of 10 hours. Limiting the retraining cycles to just once per hour can significantly boost the device's lifetime to $\approx10,000$ hours. The authors also claim that the proposed VEGA processor can reach $20\times$ the NUCLEO-64's battery for a learning event rate of one. For the second experiment, the authors initially compare the average power consumption of the smartphone used (Snapdragon 845) by Pelegrini \etal \citep{pellegrini2020latent} with the proposed processor where VEGA indicates a 9.7$\times$ higher efficiency. Afterwards, they assess the energy performance of VEGA over a minibatch of Core50 every minute, where only the fully-connected/linear layer is trained to perform inference each second. This operation consumes 0.25 Joules per minute, giving it a lifetime of 108 days.} 

To extend the widespread potential of TinyML, Grau \etal \cite{grau2021tinyfed} and Kopparapu \etal \cite{kopparapu2021tinyfedtl} demonstrate the applicability of Distributed Machine Learning concepts in tandem with TinyML (refer Section \ref{networkTinyML}). However, both implementations resort to Eager Learning approaches to improve local models. Grau, for instance, developed a Multi-class classification Key-Word Spotter. Training on a singular device, it was observed that the loss gradually degrades although the accuracy can fluctuate. Similarly, Kopparapu implemented Federated Transfer Learning (TinyFedTL) in a low resource setting, where they considered a binary classification problem (cat vs no-cat and dog vs no-dog) where the \textit{cat vs no-cat} was treated as the Transfer Learning problem. \new{Performance details of these two approaches are discussed further in Section \ref{distributedTML}}

However, the systems mentioned earlier in Eager Learning TinyML implicitly assume that training data is balanced and temporally independent. But, it can be seen that real-world data hardly comply with the stated. Taking this imbalance and temporally correlated data into account \cite{sudharsan2021imbal} devised Imbal-OL -- an online learning plugin for balancing a data stream according to the available classes prior to feeding it to the learner algorithm. By doing so, the authors effectively delay the effects of Catastrophic Forgetting (discussed further in Section \ref{catastrophic-forgetting})

\subsection{Network Reliant Approaches} \label{networkTinyML}
IoT, by definition, is a \textit{network} of interconnected objects or things. TinyML is expected to leverage the network for broader, connection-oriented applications in perseverance to extend the existing network. This classification of TinyML applications can be separated into Distributed TinyML mechanisms and Over-the-air approaches.

\subsubsection{\textbf{Distributed TinyML}} \label{distributedTML}
The pitfalls of Cloud Computing in terms of data privacy and security led to the advent of Distributed Intelligence mechanism that is centered on training on \textit{decentralized} data (E.g., Data on our mobile phones). Federated learning (FL) \cite{mcmahan2017communication} is one of the distributed ML mechanisms that focus on the concept of clients training their models at the edge without transmitting raw data \cite{imteaj2021flsurver}. The process is synonymous with a central server that occasionally aggregates local models to update a global model. FL has shown rapid growth primarily due to the strong privacy guarantees of the approach; since data remains at the edge.

The on-device learning components by Grau \etal \cite{grau2021tinyfed} and Kopparapu \etal \cite{kopparapu2021tinyfedtl} are discussed in detail in Section \ref{eagerLearning}. Kopparapu \etal \cite{kopparapu2021tinyfedtl}  laid the stepping stones in Distributed TinyML, specifically FL. It was noticed that validation accuracy depreciates with the increased number of devices. It was also observed that the Transfer learning models reached local optima after $\approx3000$ data points. Similarly, Grau \etal \cite{grau2021tinyfed} deployed FL for Independent and Identically Distributed (IID) data and non-IID data. The results show that the loss degrades over training epochs with IID data. As a result of the parameter server averaging the characteristics, it can be observed that with non-IID data, the loss increases every ten epochs.

Sudharsan \etal deviated from conventional FL implementations to provide the Globe2Train framework \cite{sudharsan2021globe2train} that allows geographically separated IoT devices to contribute to training a single model. The sparsity of GPUs compared to IoT devices drives the concept of Globe2Train. For instance, the quantity ratio between the two device types provides greater processing power as a collective compared to standalone GPUs. The ability to provide more significant resources as a collective for little to no cost incentivizes the said framework. The proposed system is composed of 2 components, i.e., G2T-Cloud and G2T-Device, that work in harmony to facilitate collaborative training.

\subsubsection{\textbf{Over-the-Air Approaches}} \label{networkTinyML}
Over-the-Air (OTA) updates are commonplace in resource-abundant situations. A driving factor of OTA updates is the availability of re-writeable secondary storage (E.g., Solid State storage, Hard drives, SD cards). Today, a majority of the MCUs come equipped with flash storage. Opposed to other secondary storage technology, the only foreseeable limitation in flash storage is the number of write cycles ($\approx100000$ cycles) \cite{kawaguchi1995flash}. However, there have been attempts to democratize TinyML in an OTA fashion.

Realizing the advantages of proliferating Embedded ML, Doyu \etal \cite{doyu2020bringing} conceptualized TinyML as-a-service (TinyMLaaS) -- an on-demand distribution framework for TinyML. This system features a cloud or web-based service with popular ML compilers whose sole purpose is to convert and distribute a specifically requested device-compatible ML model. A potential flaw with this design is the need for a constant connection to a network. Providing so will negatively affect an MCUs lifetime and make it vulnerable to security and privacy breaches.

\new{Conceptual frameworks like TinyMLaaS have yet to be implemented for a broader audience. These concepts require a strategy to transmit models from a said Cloud server to an MCU. Sudarshan \etal devised an OTA component that can supplement TinyMLaaS. They created OTA-TinyML \citep{sudharsan2022ota}, a hardware-friendly approach to repurpose MCUs running TinyML by enabling them to download and store models from the cloud. The system composes of two components. Firstly, a cloud server that provides the models on demand. Secondly, the authors include methods that enable these frugal devices to store/load the retrieved models to/from either the internal memory or external file systems. They test the solutions on 7 MCUs with different resources to ensure that the implementation works as expected for the relevant device.}

In \cite{fota2021badawy}, Badawy \etal introduced Flashing Over The Air (FOTA) -- a system to flash ATMEL AVR MCUs wirelessly. The said is achieved via Wifi and Long Range (LoRa) technology. Concretely, FOTA will permit TinyML to be updated dynamically. Unlike the solutions explained in Sections \ref{adaptiveLearning}, \ref{distributedTML} where model updates were done in the RAM,  volatility of the primary memory does not affect the flashed-solution.

\begin{table*}
\caption{\new{Comparing the state-of-the-art in reformable TinyML}} \label{tab:sota}
\centering
\footnotesize
\scalebox{0.82} {
\begin{tabular}{cccccc}
\toprule
Publication & Year & Architecture(s) & Classification & Use-case & Specialization\\

\midrule
\cite{sudharsan2020edge2train} & 2020 & SVM & \makecell{On-device \\ Offline training} & \makecell{Image Classification} & \begin{minipage}[c]{5cm}
    \begin{enumerate}
        \item First on-device offline training solutions.
    \end{enumerate}
  \end{minipage}\\
  
\midrule
\cite{disabato2020ondevice} & 2020 & $k$NN & Lazy Learning & \makecell{Keyword Spotting \\ Image Classification} & \begin{minipage}[c]{5cm}
    \begin{enumerate}
        \item Non-parametric classifier. 
        \item First on-device learning algorithm in an MCU setting.
    \end{enumerate}
  \end{minipage}\\
\midrule

\cite{cai2020tinytl} & 2020 & NN & Eager Learning & \makecell{Image Classification} & \begin{minipage}[c]{5cm}
    \begin{enumerate}
        \item Freezes the weights and updates the bias'.
        \item Done by a resource-efficient bias module named Lite Residual Module.
    \end{enumerate}
  \end{minipage}\\
\midrule

\cite{ren2021tinyol} & 2020 & NN & Eager Learning & \makecell{Anomaly Classification \\ Anomaly Detection} & \begin{minipage}[c]{5cm}
    \begin{enumerate}
        \item Insertable final yet trainable layer.
    \end{enumerate}
  \end{minipage}\\
\midrule

\cite{ravaglia2021tinyml} & 2020 & CNN & Eager Learning & \makecell{Image Classification} & \begin{minipage}[c]{5cm}
    \begin{enumerate}
        \item Continual Learning on a PULP processor.
        \item Quantization process based on Latent Replay rehearsing for reducing memory usage.
    \end{enumerate}
  \end{minipage}\\
\midrule

\cite{sudharsan2021mlmcu} & 2021 & \makecell{GD \\ SGD} & \makecell{On-device \\ Offline training} & \makecell{Image Classification \\ Medical Diagnosis} & \begin{minipage}[c]{5cm}
    \begin{enumerate}
        \item Introduced two algorithms to train TinyML models on-device. Opt-SGD to train binary classifiers and Opt-OVO to train multi-class classifiers.
    \end{enumerate}
  \end{minipage}\\
\midrule

\cite{kopparapu2021tinyfedtl} & 2021 & \makecell{Transfer Learning \\ Federated Learning} & \makecell{Eager Learning \\ Distributed TinyML} & \makecell{Image Classification} & \begin{minipage}[c]{5cm}
    \begin{enumerate}
        \item First implementation of Federated Learning in TinyML.
    \end{enumerate}
  \end{minipage}\\
\midrule

\cite{grau2021tinyfed} & 2021 & \makecell{CNN \\ Federated Learning} & \makecell{Eager Learning \\ Distributed TinyML} & \makecell{Keyword Spotting} & \begin{minipage}[c]{5cm}
    \begin{enumerate}
        \item Deployed FL for IID and Non-IID setting.
    \end{enumerate}
  \end{minipage}\\
\midrule

\cite{sudharsan2021trainplusplus} & 2021 & \makecell{non-NN} & \makecell{Eager Learning} & \makecell{Image Classification} & \begin{minipage}[c]{5cm}
    \begin{enumerate}
        \item First non-NN Image Classifier capable of online learning.
    \end{enumerate}
  \end{minipage}\\
\midrule

\cite{sudharsan2021imbal} & 2021 & \makecell{Rule-based} & \makecell{Eager Learning} & \makecell{Image Classification} & \begin{minipage}[c]{5cm}
    \begin{enumerate}
        \item A lightweight plugin to balance the class sizes in imbalanced data streams.
    \end{enumerate}
  \end{minipage}\\
\midrule

\cite{doyu2021tinymlaas} & 2021 & \makecell{TinyMLaaS} & \makecell{Over-the-Air \\ ML-as-a-Service} & \makecell{Multiple use-cases} & \begin{minipage}[c]{5cm}
    \begin{enumerate}
        \item Conceptual framework to provide TinyML models on demand based on device factors.
    \end{enumerate}
  \end{minipage}\\
\midrule

\cite{sudharsan2021globe2train} & 2021 & \makecell{Rule-based} & \makecell{Distributed TinyML} & \makecell{Image Classification} & \begin{minipage}[c]{5cm}
    \begin{enumerate}
        \item Implements a framework that leverages idle IoT resources to train networks on a global scale
    \end{enumerate}
  \end{minipage}\\
\midrule

\cite{fota2021badawy} & 2021 & \makecell{Flashing OTA} & \makecell{Over-the-Air} & \makecell{N/A} & \begin{minipage}[c]{5cm}
    \begin{enumerate}
        \item Approach to flash an MCU OTA via Wifi and LoRa.
    \end{enumerate}
  \end{minipage}\\
\midrule

\cite{sudharsan2022ota} & 2022 & \makecell{OTA TinyML} & \makecell{Over-the-Air} & \makecell{Image Classification \\ Audio Classification \\ Anomaly Detection} & \begin{minipage}[c]{5cm}
    \begin{enumerate}
        \item Approach to load and execute TinyML models from the cloud in a hardware efficient manner.
    \end{enumerate}
  \end{minipage}\\
\bottomrule

\end{tabular}}
\end{table*}

\begin{table*}
\caption{\new{Software \& Hardware analysis of the state-of-the-art in reformable TinyML (Legend -- \textbf{P}: Publication, \textbf{OnR}: Online Resources, \textbf{SW}: Software Leveraged)}} \label{tab:sota-performance}
\centering
\small
\scalebox{0.82} {
\begin{tabular}{cccccc}
\toprule
P & OnR & SW & Devices & Dataset(s) & Evaluation metrics\\
\midrule

\cite{sudharsan2020edge2train} & \cmark & \makecell{C++} & \makecell{Adafruit Feather nRF52840 \\ STM32F103C8T6 \\ ESP32-S2 \\ Adafruit HUZZAH32 \\ Adafruit METRO M0} & \makecell{Iris Flowers \\ Breast Cancer \\ MNIST} & \makecell{Training time \\ Inference time \\ Energy consumption} \\
\midrule

\cite{disabato2020ondevice} & \xmark & \makecell{C \\ Python} & \makecell{Raspberry Pi 3B+ \\ STM32F7} & \makecell{Speech Commands Dataset \\ Imagenet} & \makecell{Accuracy \\ Memory usage \\ Inference time} \\
\midrule

\cite{cai2020tinytl} & \cmark & \makecell{Python} & \makecell{Arduino Nano 33 BLE} & \makecell{Imagenet \\ Visual Wake Words} & \makecell{Accuracy \\ Memory usage \\ Total training time} \\
\midrule

\cite{ren2021tinyol} & \xmark & \makecell{C++} & \makecell{Arduino Nano 33 BLE} & Custom dataset & \makecell{Mean Squared Error \\ Inference time \\ F1-score \\ Macro F1-score} \\
\midrule

\cite{ravaglia2021tinyml} & \xmark & \makecell{Pytorch \\ Python} & \makecell{VEGA \\ STM32L476RG} & \makecell{CORE50} & \makecell{Accuracy \\ Memory usage \\ Latency \\ Energy Efficiency \\ Hardware/Software Efficiency} \\
\midrule

\cite{sudharsan2021mlmcu} & \cmark & \makecell{C++} & \makecell{Adafruit Feather nRF52840 \\ STM32F103C8T6 \\ ESP32-S2 \\ Adafruit METRO M0} & \makecell{Iris Flowers \\ Breast Cancer \\ MNIST \\ Australian Sign Language \\ Heart Disease} & \makecell{Accuracy \\ Training time \\ Inference time \\ Memory usage \\ Storage usage} \\
\midrule

\cite{kopparapu2021tinyfedtl} & \cmark & \makecell{C++ \\ Python} & \makecell{Arduino Nano 33 BLE} & \makecell{Imagenet \\ Visual Wake Words} & \makecell{Accuracy \\ Memory usage \\ Total training time} \\
\midrule

\cite{grau2021tinyfed} & \xmark & \makecell{C++} & \makecell{Arduino Nano 33 BLE} & \makecell{Custom KWS datasets} & \makecell{Accuracy} \\
\midrule

\cite{sudharsan2021trainplusplus} & \cmark & \makecell{C++} & \makecell{Adafruit Feather nRF52840 \\ STM32F103C8T6 \\ ESP32-S2 \\ Adafruit HUZZAH32 \\ Adafruit METRO M0} & \makecell{Iris Flowers \\ Breast Cancer \\ MNIST \\  Banknote Authentication  \\ Heart Disease \\ Haberman's Survival \\ Titanic} & \makecell{Accuracy \\ Training time \\ Energy consumption} \\
\midrule

\cite{sudharsan2021imbal} & \xmark & \makecell{C++} & \makecell{Google Coral \\ Raspberry Pi 4 \\ NVIDIA Jetson Nano} & \makecell{CIFAR10 \\ CIFAR100} & \makecell{Accuracy \\ Memory usage \\ Elapsed time} \\
\midrule

\cite{doyu2021tinymlaas} & \xmark & N/A & \makecell{N/A} & \makecell{N/A} & \makecell{N/A} \\
\midrule

\cite{sudharsan2021globe2train} & \xmark & N/A & \makecell{Adafruit Feather nRF52840 \\ STM32F103C8T6 \\ Adafruit METRO M0} & \makecell{ Australian Sign Language} & \makecell{Accuracy \\ Training time} \\
\midrule

\cite{fota2021badawy} & \xmark & C & \makecell{Arduino MEGA ADK \\ Arduino UNO} & \makecell{N/A} & \makecell{N/A} \\
\midrule

\cite{sudharsan2022ota} & \cmark & C++ & \makecell{Teensy 4.0 \\ STM32 Nucleo H7 \\ Arduino Portenta \\ Feather M4 Express \\ ESP32 \\ Arduino Nano 33 \\ Raspberry Pi Pico} & \makecell{N/A} & \makecell{Memory usage} \\

\bottomrule

\end{tabular}}
\end{table*}

\subsection{\new{Analysis of the Software implementations}}

\new{Analyzing the software implementations of reformable systems is imperative to understand benchmarking, evaluation and future improvements. Table \ref{tab:sota-performance} summarizes the software leveraged, Hardware selections, datasets used, availability of resources for reproducibility and evaluation metrics. In terms of software, it is evident that C++ is the predominant language selected for reformable TinyML solutions. As C++ is energy efficient, stable, and sits closer to the device's hardware, it has been the go-to language for developing these solutions and pre-existing embedded systems. This means that developing with C++ increases the portability of implementation, thus, making it easier to test and evaluate on another commodity MCU. By looking at the evaluation metrics used, it can be seen that researchers have resorted to different metrics in different solutions. However, since many solutions are targeted at classification problems, the metrics Accuracy, Memory usage and Inference Time are observed recursively.}

\subsection{Benchmarking Frameworks for TinyML} \label{benchmarking}

As previously mentioned, the multitude of available MCUs in the market makes benchmarking in TinyML inevitably arduous \cite{banbury2020benchmarking}. Concretely, the lack of benchmarking suites for ultra low powered MCUs disrupts the momentum gained by TinyML. To fill this gap, Banbury \etal \cite{banbury2021mlperf} and Sudarshan \etal \cite{sudharsan2021benchmark} introduced benchmarking suites to propel further research in TinyML. 

Banbury \etal defined the TinyMLPerf suite, consisting of 4 benchmarks, namely, Keyword spotting, Visual Wake Words, Image Classification, and Anomaly Detection. They consider energy usage,
inference latency, hardware accuracy, runtimes, and ML models as assessing metrics. Moreover, to accommodate flexibility and all-inclusiveness, the suite has two divisions; A flexible Open division and a restricted Closed division for submitting results. The Closed division focuses on comparing the inference stack, i.e., the runtime, SoC, compiler, etc., whereas the Open division centers around the model architecture. In order to maintain stability among the results, they also provide reference models for each of the benchmarks that run on the TFLM runtime. Similarly, Sudarshan \etal in the steps of Banbury \etal Benchmarked seven popular MCUs with ten datasets. The experiment was conducted for three types of fully connected NNs, summing the analysis up to a total of 30 NNs. Similar to Banbury, TFLM was chosen as the underlying TinyML runtime. The summaries of both benchmarking suites are tabulated in Table \ref{tab:benchmarking}

\begin{table*}
\caption{Summary of benchmarking suites in TinyML} \label{tab:benchmarking}
\centering
\small
\scalebox{0.85} {
\begin{tabular}{ccccc}
\toprule
Suite & Usecase  & Dataset & Reference Model(s) & Metrics\\
\midrule
\cite{banbury2021mlperf}  & 
\makecell{Keyword spotting \\ Visual Wake Words \\ Image Classification \\ Anomaly detection} & 
\makecell{Speech Commands \\ VWW Dataset \\ Cifar10 \\ ToyADMOS } &
\makecell{DS-CNN \\ MobileNetV1 \\ ResNet \\ FC-AutoEncoder} & \makecell{Latency \\ Accuracy \\ Energy} \\

\midrule
\cite{sudharsan2021benchmark}  & 
\makecell{Audio classification \\ Image Classification \\ Wave classisfication} & 
\makecell{Iris Flowers \\ Wine \\ Japanese Vowels \\ Vehicle Silhouettes \\ Anuran Calls \\ Breast Cancer Wisconsin\\ Describable Texture \\ Sensorless Drive Diagnosis \\ MNIST \\ Human Activity} &
\makecell{1 layer with 10 Neurons \\ 10, 50 neurons for 1st, 2nd layers \\ 10, 10 neurons for 1st, 2nd layers} & \makecell{Inference Time \\ Model size \\ Compile time of Arduino IDE \\ SRAM memory} \\

\bottomrule
\end{tabular}}
\end{table*}
\section{Potential of Reformable TinyML in Industrial applications} \label{potential}

Many existing IoT applications have been fused in our daily activities, extensively utilizing networking to provide us with "smart" applications. As with the miniaturization of IoT, prior to TinyML, applications falling under the Extreme Edge criteria were typically dumb and limited to transferring data to and from a parent node (E.g., Smartphone). For instance, wearables, i.e., external accessories that can be worn (E.g., smartwatches, smart glasses), have seen widespread progression during the previous years \cite{dian2020wearables}. However, seeing these devices linked to a parent node to exploit its potential is still common. The advent of TinyML has made it possible for low-resourced devices to deviate from this interdependence. The untethering of low-resource devices from a parent node allows for previously untapped applications. Out of the many areas and applications that show potential, we have filtered out a few that will benefit the most out of reformability.

\subsection{Industrial IoT}
Recently, Industrial applications have seen the amalgamation of "smart objects" with mechanical or automated components. Referred to as Industry 4.0, it envelops the central concept of smart-manufacturing with the inclusion of cyber-physical systems enabled by IoT \cite{frank2019industry, lasi2014industry}. 

With extensive quantities of data generated by Industrial IoT, Tiny ML in particular, can  help benefit various industries.
\new{One key application area would be to monitor/diagnose entire industrial complexes with various calibres of machinery using signal classification and predictive maintenance models. For instance, consider the example of an industrial complex which hosts a production line to build commodities. In such cases, a malfunction or machinery failure can reduce the production rate or worse; it can halt the production line completely. The after-effects of such an event would require the commodity manufacturer to incur the costs for urgent replacement parts and repairs for the failed machine on top of losses resulting from the reduction in production rate. A remedial measure to mitigate such misfunctions would be to monitor crucial machinery with always-on ML, i.e., TinyML, to address any anomalies before they escalate.}

\new{Considering batch learning, it involves the need for precompiled datasets from which TinyML models can learn. This dataset would ideally be compiled by capturing events/misfunctions previously encountered and recorded. Unless otherwise the data has been recorded, it will be strenuous to replicate it to robustify the predictive maintenance model. Furthermore, in the context of MCUs, storing sensor data may not be feasible for future extraction because most devices have deficient storage capacities. Even given that data could be saved internally, data extraction can be a cumbersome task based on the placement of the device, i.e. if the devices are embedded closer to core components, delicate procedures might be imperative.}

\new{We must also consider that, in reality, even identical machines can encounter unpredictable and contrasting misfunctions. By extending the example above, let us consider that $A_1$ and $A_2$ are different anomalies identified in two identical devices $D_1$ and $D_2$ at times $T_n$ and $T_{n+m}$. Assuming $D_1$'s adaptation to $A_1$ has been implemented in $D_2$, it would be a practical choice to accommodate the solution identified for $A_2$ to $D_2$ for future-proofing purposes. For batch learning networks, it would involve extracting data from all the MCUs, compiling a collective dataset, retraining and deploying. The labour required to complete the pipeline can be substantial based on the location of the MCU in its respective machine. In retrospect, an Online Learning approach can be implemented such that embedded devices can improve the models by themselves. Later, a Federated TinyML can be introduced later such that devices in several machines can collaboratively improve a centralized ML/DL model; as a result, improving all ML/DL models in each of the MCUs.}

\subsection{Healthcare}

 Healthcare is a sector that sees excellent growth potential with abundant smart frugal devices. Today, wearables dominate the market for affordable health monitoring. However, caveats of existing solutions are the reliance on a primary device (E.g., smartphone) and the significant energy budget to perceive, transfer data and receive predictions in a Wireless Sensor Network (WSN); hence, why many devices are rechargeable. \new{The reliance on a primary device to transfer to the fog/cloud can impact inference speeds since processing could require data to be telemetered outside an MCUs local network. This can instigate delays that can be potentially life-threatening to a monitored patient, as mentioned by Bharadwaj \etal \cite{bharadwaj2021review}.}  In addition to these, the sensitivity of health data inherently requires measures to ensure sufficient privacy, especially in a WSN, a quality that the limited resources could hinder in wearables. 
Alternatively, TinyML can localize the training and processing of sensor data, allowing for novel untethered applications that can be used for extended periods without the need for regular charging. \new{For example, activity monitors have grown in capacity and utility with the advancements in wearables for recreational purposes. A trait of such devices would be that they are personal devices often used by a single individual throughout their lifetime. These individuals have patterns that vary from one another and, most certainly, vary from a wearable's default settings/parameters. An individual who regularly jogs will have a different heart rate than someone who speedwalks. Identifying and adapting to these patterns of an individual over time and age can generate better exercise-related statistics for a given individual. For personalization purposes, incremental learning can be leveraged. FL can be implemented in instances where sensitive data such as heart rate and blood sugar levels can be used to help research and improve predictive tools for a general/broad population.}

Thus, TinyML can completely strip the need for networking in eHealth solutions or work in sync with a central server to provide FL implementations that can thrust eHealth to its full potential.

\subsection{Smart Environments/Cities}
The integration of ML-driven extreme IoT will permit ubiquitous intelligence alongside supplementing present-day IoT in domains such as security, smart vehicles, moving things \cite{prado2021robustifying}, traffic monitoring and mitigation \cite{roshan2021adaptive}, Wildlife conservation, and pavement anomaly detection \cite{andrade2021pavement} among many things. In addition to this, the decentralized nature of TinyML provides the capacity and responsiveness to make near-instantaneous predictions which are vital for delay-critical tasks such as traffic control. 

\subsubsection{\textbf{Smart Environments}}
In the current context, the ability to execute inexpensive ML tasks deprived of a continuous power source provides the added advantage of disregarding the need to telemeter data to and from an MCU. This is especially beneficial in network-less contexts such as rural areas, creating "smart" environments that aid the rural population in their day-to-day. In such scenarios, the inability to provide OTA updates, in the long run, can hinder the disadvantaged community's progression, making online learning applications prospective. \new{For instance, a persistent problem that can be observed today, especially in rural areas, is the human-wildlife conflict. As the human population explodes, rural areas gradually expand beyond the borders of the wilderness. With it, humans and animals are constantly seen clashing with one another, where wildlife is hurt or killed most often than not. This entails adverse effects on animal populations that can harm an ecosystem's balance. As much as it affects the progression of a rural community, the conflict also affects the progression of wildlife conservation efforts. As an energy-efficient, inexpensive and off-the-shelf alternative to smartphones, dedicated cameras or drones, TinyML can assist in wildlife conservation efforts similar to what has already been done by \citep{olsson2022conservation}. In the field, image classification parameters such as background, illumination, and noise, to name a few, can vary from the original training set, given the geographical location. Referred to as Data Nonstationarity \cite{mai2022online}, incrementally adapting to these variations each time a sample is received can improve the network's performance, helping conservation efforts.}

\subsubsection{\textbf{Smart Spaces/Cities}}

\new{On the flip side, the United Nations \cite{owidurbanization} mentions that by 2050, close to 7 Billion people will reside in urban areas. With infrastructure and resources in cities being strained already, introducing information and technology can alleviate much of the existing burden \cite{ismagilova2019smart}. For instance, the energy demands for cities grow exponentially with the increase in population. As a result, it becomes even more challenging to reach strict limits set on reaching carbon neutrality, making it crucial to reduce energy wastage as much as is viable. Visual wake words \cite{chowdhery2019visual} show potential candidacy in this cause as it can be leveraged to toggle a light on/off based on the presence or absence of a person in the perspective of a TinyML-driven camera. Such implementations localized to a specific space in a house/building can be expanded further, effectively forming smart buildings where communication between sensors and consistent power can be facilitated. Expanding this concept further, smart cities can eventually be formed by constructing several sensor-rich smart buildings to reduce the excessive utilization of resources. With stable networks formed by other interconnected devices and appliances in the smart spaces, TinyML can resort to OTA updates by leveraging these pre-existing networks to improve its collective performance, thereby increasing its city's efficiency even further. Similarly, TinyML can be vastly integrated into smart spaces that will allow users to receive services by interacting with various sensors and actuators in a user's physical environment \cite{soro2021tinyml}.}

\subsection{Agriculture}

Today, many nations rely on crops for sustenance and capital, making monitoring of crops indispensable. \new{However, crop diseases are the most significant blocker in high-yield crops, resulting in many economic and production losses.} Smart agriculture, as mentioned by \cite{gondchawar2016iot}, is the solution to various inhibitors of agricultural output. Including smart monitoring tools in farms and fields can increase production with higher quality, which translates to higher revenue. Furthermore, thanks to the TinyML revolution, inexpensive ML hardware for monitoring is available to all. However, Seasonality is a potential inhibitor of traditional diagnosis models in agriculture \new{, i.e., once the model parameters have been established, it can no longer adapt to its surrounding environment. Incremental/online learning, as mentioned in \cite{zhu2013incrementalagri}, can benefit in terms of adaptations to temporal changes in crops.} Given the vastness of agricultural areas, network-reliant approaches such as federated learning can also be adapted to benefit the broader collective of diagnosis models. Given the frugality of MCUs, communication means with Long-Range (LoRa) will have to be adopted for such approaches to compensate for weak RATs available in most devices.

\subsection{Augmented/Virtual Reality}
Augmented Reality (AR) and Virtual Reality (VR) are exponentially growing technologies focused on enhancing user experience by immersing them in an entirely virtual (VR) or semi-virtual (AR) interactive space. As said in \cite{khan2019ar}, AR has three requirements that need to be fulfilled: 1) The ability to combine real and virtual objects in a physical space. 2) The ability to align real and virtual objects with one another. 3) The ability to interact with the AR space in real-time via gestures, motion-tracking, etc. Contemporary retail gesture-based interactive products for AR/VR are often a luxury that inhibits the widespread use of such systems. 

Provided this, the research aims to drastically reduce the cost of these systems by introducing frugal devices with sufficient processing capabilities as cheap alternatives to expensive hardware. Furthermore, the dependence of existing research on a resource-rich device for processing has given TinyML a competitive advantage. For instance, Bian \etal \cite{bian2021capacitive} introduced a wrist-worn gesture-recognition system that processes the capacitor data locally by leveraging TinyML, reducing induced latency of processing interactions off-device. Gestures or movements vary from individual to individual, making a system similar to Bian \etal prone to false predictions due to the use of a static model trained offline. Concretely, the capability to learning on-the-fly \new{using online learning}, benefits by allowing the classifier to classify varying gestures under a particular class. Additionally, the users will benefit from the model personalizing to fit their movements/gestures. 
\section{Open challenges, Future directions and Next generation TinyML} \label{challengesFutureTinyML}

Despite the growth of TinyML, the new domain faces challenges that, as a result, open new research pathways. This section highlights the challenging problems and the future research directions in this field.

\subsection{Challenges}

\subsubsection{\textbf{Catastrophic Forgetting}} \label{catastrophic-forgetting}
Inspired by biology, adaptations to Concept drift have allowed ML to adapt to changing environments indefinitely. However, unlike biological counterparts, NNs leveraging online learning techniques to train on multiple tasks are prone to \textit{Catastrophic Forgetting}. Catastrophic Forgetting is a phenomenon commonly seen in adaptive algorithms where the performance abruptly decreases when new classes are used to train an existing model. This, as mentioned by \cite{kirkpatrick2017catastrophic}, is a result of having to change the weights required for task A to fit the requirements of a new task B. Although online-TinyML capabilities can supplement existing work in IoT, in this context, there is a considerable challenge to overcome Catastrophic Forgetting given the resource scarcity.

\subsubsection{\new{\textbf{Benchmarking}}} \label{benchmarking-issue}
\new{TinyML, as it is, is already plagued by the need for proper benchmarks. As mentioned in Section \ref{heterogeniety}, this is driven by the variety of devices that host their scarce resources. Reformability in TinyML escalates the said issue even further. Once again, the performance of a solution is inherently dependent on the resources an MCU has. For example, a standardised implementation as a benchmark for a device may not be implementable in a device with lesser resources. Considering many devices and configurations exist, evaluation can be subjective to the device used, i.e., the performance metrics vary from device to device. This makes benchmarking the comparatively resource-heavy reformable TinyML solutions excruciating and uneasy.}

\subsubsection{\textbf{Resource constraints}}
Standard ML solutions executed in the cloud or workstations often do not have limitations for resource utilization. On the contrary, frugal devices, a subset of edge devices, are orders of magnitude more constrained. Most MCUs are accompanied by memory measured in Kilobytes instead of Megabytes and processors in the MegaHertz instead of GigaHertz in the newest mobile phones. However, the lack of resources is TinyMLs biggest dilemma. To recall, lack of resources help achieve great efficiencies while limiting computational capabilities. In the line of on-device learning, this is a major inhibitor. As mentioned in Section \ref{catastrophic-forgetting}, Catastrophic Forgetting plagues online learning implementations. For instance, according to \cite{hayes2020remind}, replay, i.e., fine-tuning a NN with data points from old and new tasks, is a common remedy. However, accommodating old data points for replaying requires storage/memory; and doing so in the limited storage/memory in MCUs is improbable. In this regard, we emphasize the need for constraint-friendly online learning solutions to reach TinyML's full potential.

\subsubsection{\textbf{\new{Infancy and generalizability}}}
\new{With TinyML's potential to exploit inexpensive hardware for ML/DL, we can see the advent of various tools to ease the development process. Table \ref{tab:code_generation} summarizes the available tool sets that can be leveraged to deploy TinyML solutions in mere minutes. However, all the available tool sets are specific to static TinyML, i.e., these tools do not support reformability. This is a direct result of reformable TinyML's infancy. For instance, Static TinyML has built a reputation regarding reliability with several real-world implementations, such as Keyword Spotting for Smart assistants. The same cannot be said about reformable TinyML, as it has yet to prove or exceed expectations in reliability in real-world applications.}

\new{Furthermore, generalizing reformable TinyML solutions is another hurdle that needs overcoming. These works have higher resource overheads making them mostly implementable on devices with contextually higher resources, e.g., Arduino Nano 33 BLE. Making the existing work universal for all devices, especially devices with lower resources, may require the implementations to be modified drastically, adversely affecting performance in the real world.}

\subsubsection{\textbf{Network management}}
The supplementary nature of many TinyML solutions in IoT ecosystems inherently focuses on utilizing Radio Access Technologies (RATs) to extend functionality. Given that MCUs typically run on batteries, networking components in MCUs require substantially higher power to function than processing on-premise. For context, \cite{reijers2003efficient} states that the transmission of 1 bit has the energy consumption equivalent to executing 1000 instructions onboard. Accordingly, it is vital to manage network resources carefully and sparingly to ensure device longevity. For this purpose, TinyML can also assist in intelligent communication where reducing the usage of RATs is a priority. Further, algorithms similar to that in \cite{sanchez2020tinyml} can be utilized for choosing the optimal RAT in multi-RAT systems to preserve battery. \new{Alternatively, solutions similar to ElastiCL \cite{sudharsan2021elasticl} can be used to decrease the number of transmissions needed by dynamically altering the quantity of bits in a model.} For delay-tolerant approaches, task-scheduling can be considered to inhibit the use of RATs unless scheduled.

However, it is worth mentioning that the placement of MCUs can and are capable of functioning in drastically varying environments (E.g., Implanted devices, Machine-embedded devices) in various network contexts. Adapting to the different network contexts is another research area that will need consideration with network-reliant TinyML to provide MCUs with a certain degree of autonomy.

\subsubsection{\textbf{Volatility of the Primary memory}}
As a component of resource constraints, the primary memory (SRAM) of MCUs ranges between a few to a few hundred KiloBytes. As mentioned by \cite{ren2021tinyol, warden2019tinyml}, NNs incorporated into a flash as a C/C++ array are treated as frozen graphs, meaning that modifications/updates to this graph are prohibited. Thus, many online learning solutions trains/improve the existing model as a whole \cite{kopparapu2021tinyfedtl} or partially \cite{ren2021tinyol} in the SRAM without saving it in the flash memory. The volatility of the SRAM means that progress made in training a model will be erased when the MCU is reset or powered down. To mitigate the said issue, concepts from Flashing Over-the-Air \cite{fota2021badawy} could be applied to ensure that progress is saved.

\subsubsection{\new{\textbf{Cyber security}}} 
\new{Many concurrent TinyML solutions hardly include a networking component as processing sensor data is completed within the MCU. However, cyber security is a major concern for TinyML solutions that leverage RATs to communicate with external devices, e.g., Federated TinyML. Furthermore, Gill \etal state that Edge devices, including MCUs, are built for simplicity and to be inexpensive, not for security \citep{gill2022ai}. In the context of TinyML, security issues are once again coupled with an MCUs resource scarcity. Running robust security software and firewalls is not viable for these frugal devices, making assessing and enabling resource-friendly security features imperative.}

\subsubsection{\new{\textbf{Physical security}}}

\new{In Edge computing, it is vital to have some degree of control over physical access to the devices to ensure security. In the context of TinyML, a device's size and lack of resources can make it difficult to restrict access and provide adequate physical security. Furthermore, when paired with device heterogeneity, the ability to provide security varies significantly from device to device. For example, Indoor localization can be used to locate or prevent the theft of an MCU in an area. However, it requires a RAT to relay its current location/position, meaning that devices void of RATs is vulnerable to theft. Thus, generalizing physical security for all calibres of MCUs is a persistent challenge.}

\subsubsection{\new{\textbf{Energy Evaluations}}}
\new{It was stated in Section \ref{benchmarking-issue} that device heterogeneity makes it a considerable challenge to generalize benchmarking across all MCU platforms. The lack of a standardized benchmark to assess a reformable model's performance inherently causes the lack of common energy evaluations that can be used to compare one another. As extreme energy efficiency is core to TinyML, without a common energy assessment, the ability to select the most suitable technique out of the available techniques for a specific use case diminishes.}

\subsection{Future Directions}
\subsubsection{\textbf{Task offloading}} \label{offloading} 

Task offloading is the transfer of computational responsibility to more resource-rich devices such as Edge, Fog, or Cloud computing services \cite{saeik2021task}. 
Encompassing the limitations of embedded devices and existing work to formulate effective Online learning solutions, offloading learning/training, or other intensive processing tasks to resource-rich devices needs investigation. A Cyber Foraging add-on is favorable as it benefits the frugal devices in quick and energy-efficient processing. Cyber-foraging is the exploitation of resources in \textit{static idle} devices (referred to as "surrogate" devices) in close vicinity to a mobile device (battery-powered, portable devices) \citep{nithya2020sdcf}. In the context of TinyML, shifting cloud processing to the Fog can balance the trade-off between the required energy for radio communications and latency paired with processing resources in a Fog node.

\subsubsection{\textbf{Communication schemes}} \label{communication-schemes} MCUs have the ability to operate unattended using battery or ambient energy harvesting methods. The guarantees in longevity for these devices in network-reliant reformability revolve around routine operations such as model transfers or retrievals. Depending on the model's size, the number of messages that need to be received/transmitted vary. Consequently, it affects power requirements and the time consumed for communication tasks. Assuming the possible network contexts due to MCUs placement, reducing redundant communications and re-transmissions of data is essential. For this purpose, we suggest succinct intelligence mechanisms in the Application layer can be formulated to use contemporary RATs such as BLE or LoRa.

\subsubsection{\textbf{Data Falsity}} Denoting the accuracy of data sources, "Veracity" is one of the 5Vs that constitute Big data \cite{anuradha2015brief}. Most of the existing solutions in online learning imply that the data source is perfect, whereas actual data could be incomplete or be contaminated with noise. Without advanced preprocessing, this could hinder the performance of algorithms post-deployment. Thus, we emphasize the need for future research to address the veracity problem by robustifying online learning algorithms.

\subsection{\new{Next Generation TinyML}}

\new{At its current momentum, TinyML will undoubtedly see the union of other popular technologies. This section summarizes potential next-generation applications for TinyML that are derived from \cite{gill2022ai}.}

\subsubsection{\new{\textbf{Integration of Blockchain}}} \label{blockchain}
\new{Throughout the past few years, Blockchain has been able to retain the spotlight in discussions pertaining to cybersecurity. Blockchain is a decentralized, immutable ledger holding historical information. This immutability can help secure network-reliant reformable TinyML solutions which necessitate a network. To recall, requiring to communicate with external devices means that these MCUs will need to expose their RATs throughout the lifetime of the connection/transmission. This inherently entails risks, especially given the resource scarcity of these devices. Despite this, private blockchains like Hyperledge Fabric can be applied alongside TinyML to improve security in the deployment stage of the TinyML pipeline. These ledgers can be used to transmit software or models via secure peer-to-peer propagation, which according to Taylor \etal \citep{taylor2020systematic}, is by far one of the more widespread use cases of blockchain in the domain of IoT.}

\subsubsection{\new{\textbf{Explainable AI}}}
\new{The simplest ML/DL models are black boxes where any internal processes are incomprehensible to humans. Explainable AI (XAI) is a set of processes that allow humans to interpret the internal processes that eventually help characterize the model's accuracy, bias' and trustworthiness, thus, helping the AI community advance further. However, characterizing a model becomes complicated when a model is dynamic and constantly changes, such as in Online learning TinyML models. As mentioned by \citep{de2022perils}, the results/explanations obtained yesterday can be obsolete or drastically different today. Nevertheless, for solutions that resort to some form of OTA updates where the transmitted model is static, e.g., FL, XAI can be imperative to understand the model's performance.}  

\subsubsection{\new{\textbf{Decentralization}}}
\new{As previously mentioned, Edge computing deviates from CC to retreat from the risks of centrally processing/storing data. Apart from influencing decentralization for robustifying security (Section \ref{blockchain}), we state that it can be applied in TinyML, especially for Network reliant solutions. For instance, Blockchain can be infused with Globe2Train \cite{sudharsan2021globe2train}, and GL \cite{hegedHus2019gossip} to leverage idle resources such that a single model can be refined and distributed in a secure and decentralized manner. However, it can be stated that decentralization in TinyML can significantly increase resource usage across all devices. We would argue that considering the sheer quantities of MCUs, the available idle computing potential can be used to maximize the domain's capabilities at a low energy and resource budget.}

\subsubsection{\textbf{\new{Fusion of Serverless computing and TinyML}}}

\new{Serverless computing is a cloud computing paradigm that relieves developers from having to manage servers and their resources. Driven by its architecture to self-manage the resources of a server, the developers can spend more time and resources focusing on AI-related tasks than managing servers.}

\new{Today many businesses have on-demand applications where services are provided to the user once requested. Serverless computing sees great potential in such applications since the resources are managed based on the demand at a given time. Similarly, serverless computing can be used in TinyML with solutions like TinyMLaaS \citep{doyu2021tinymlaas} or even instances where task scheduling Collaborative TinyML is used to provide ample resources if and only if necessary, preventing unwanted energy wastage--as a result, aligning well with the TinyMLs energy efficient trait.}

\subsubsection{\textbf{\new{Advancements of ML/DL}}}

\new{ML/DL is in an era of hyper-growth, where algorithms are constantly changing to reach new heights in performance and capabilities -- each of these new works can benefit its stakeholders elsewhere. In the context of reformable TinyML, autonomy plays a considerable role. For example, apart from Online learning algorithms alone, recent algorithms such as Deep Reinforcement Learning can help drive autonomous TinyML to the next tier by learning as a human would. Likewise, the need to port other novel and popular algorithms to the domain of TinyML could boost the performance of pre-existing static or non-static TinyML implementations.}
\section{Conclusion}

Providing frugal devices with cognition widens the scope of IoT due to their packed but capable form factors and energy efficiencies. Most TinyML solutions cannot improve the embedded solution to combat an environment's dynamicity and volatility. 

This paper delineates reformable TinyML solutions, i.e., updatable solutions. In the process we
discuss the generic workflow for producing TinyML solutions where we individually describe the various methods for deploying them. We then specify the distinctive features of Embedded systems and the reformable inference mechanism applied to these devices. Promptly, we classify reformable TinyML using a new taxonomy that segregates the available literature based on the technology used to provide improvements. Here, it is worth noting that the suitability of each method is discussed in further detail. Under the classification of reformable TinyML, we discuss how the solutions improve themselves and discuss the critical components that facilitate it. With this, we also survey the limited benchmarking methods. Afterwards, we select a few industrial applications where reformable TinyML can inherently benefit the various stakeholders involved and specify how it can benefit. After that, immediately discuss the future directions. 
\new{Finally, with the advancements in computer science and software bringing new technologies, we analyze the prospects of fusing next-gen computing with TinyML.}

\section{Glossary}

\new{
\textbf{AR} - Augmented Reality \\
\textbf{CC} - Cloud Computers \\
\textbf{CNN} - Convolutional Neural Network \\
\textbf{DL} - Deep Learning \\
\textbf{FL} - Federated Learning \\
\textbf{GD} - Gradient Descent \\
\textbf{GL} - Gossip Learning \\
\textbf{IoT} - Internet of Things \\
\textbf{Kb} - Kilobytes \\ 
\textbf{KD} - Knowledge Distillation \\
\textbf{kNN} - K-Nearest Neighbor \\
\textbf{LLN} - Low power and Lossy Network \\
\textbf{MCU} - Microcontroller Unit \\
\textbf{ML} - Machine Learning \\
\textbf{NN} - Neural Network \\
\textbf{OS} - Operating System \\
\textbf{OTA} - Over the Air \\
\textbf{RAT} - Radio Access Technology \\
\textbf{RB} - Raspberry \\
\textbf{SGD} - Stochastic Gradient Descent \\
\textbf{SRAM} - Static Random Access Memory \\
\textbf{SVM} - Support Vector Machine \\
\textbf{TFLM} - TensorFlow Lite Micro \\
\textbf{TinyML} - Tiny Machine Learning \\
\textbf{VR} - Virtual Reality \\
\textbf{WSN} - Wireless Sensor Networks \\
\textbf{XAI} - Explainable Artificial Intelligence}

\bibliographystyle{ACM-Reference-Format}
\bibliography{main}

\end{document}